\documentclass[letterpaper, 10 pt, conference]{ieeeconf}  

\makeatletter
\def\endthebibliography{%
	\def\@noitemerr{\@latex@warning{Empty `thebibliography' environment}}%
	\endlist
}
\makeatother

\IEEEoverridecommandlockouts                              
\usepackage{cite}
\usepackage{xcolor}
\usepackage{graphicx}
\usepackage{amssymb}
\usepackage{amsmath}
\let\labelindent\relax
\usepackage{enumitem}
\usepackage{algorithm}
\usepackage{algpseudocode} 
\usepackage{array}
\usepackage{mathtools}
\usepackage{multirow}
\usepackage{caption}
\usepackage{subcaption}
\usepackage{booktabs}
\usepackage{hyperref}
\usepackage{cleveref}
\usepackage{url}

\renewcommand{\algorithmicrequire}{\textbf{Input:}}
\renewcommand{\algorithmicensure}{\textbf{Output:}}

\graphicspath{{./images/}}
\DeclareGraphicsExtensions{.pdf,.eps}

\title{\LARGE \bf
OCRTOC: A Cloud-Based Competition and Benchmark for Robotic Grasping and Manipulation
}

\author{Ziyuan~Liu$^1$,
        Wei~Liu$^1$,
        Yuzhe~Qin$^2$,
        Fanbo~Xiang$^2$,
        Minghao~Gou$^{1,7}$,\\
        Songyan~Xin$^3$,
        Maximo A.~Roa$^4$,
        Berk~Calli$^5$,
        Hao~Su$^2$,
        Yu~Sun$^6$,
        Ping~Tan$^{1,8}$\\\\
        $^1$Alibaba AI Labs, $^2$UC San Diego, $^3$University of Edinburgh, $^4$German Aerospace Center, $^5$Worcester\\ Polytechnic Institute, $^6$University of South Florida, $^7$Shanghai Jiaotong University, $^8$Simon Fraser University
\thanks{Ziyuan Liu is corresponding author. Email: \href{mailto:ziyuan-liu@outlook.com}{ziyuan-liu@outlook.com}}}

\begin{document}

\maketitle
\thispagestyle{empty}
\pagestyle{empty}

\begin{abstract}
In this paper, we propose a cloud-based benchmark for robotic grasping and manipulation, called the OCRTOC benchmark. The benchmark focuses on the object rearrangement problem, specifically table organization tasks. We provide a set of identical real robot setups and facilitate remote experiments of standardized table organization scenarios in varying difficulties. In this workflow, users upload their solutions to our remote server and their code is executed on the real robot setups and scored automatically. After each execution, the OCRTOC team resets the experimental setup manually. We also provide a simulation environment that researchers can use to develop and test their solutions. With the OCRTOC benchmark, we aim to lower the barrier of conducting reproducible research on robotic grasping and manipulation and accelerate progress in this field. Executing standardized scenarios on identical real robot setups allows us to quantify algorithm performances and achieve fair comparisons. Using this benchmark we held a competition in the 2020 International Conference on Intelligence Robots and Systems (IROS 2020). In total, 59 teams took part in this competition worldwide. We present the results and our observations of the 2020 competition, and discuss our adjustments and improvements for the upcoming OCRTOC 2021 competition. The homepage of the OCRTOC competition is \url{www.ocrtoc.org}, and the OCRTOC software package is available at \url{https://github.com/OCRTOC/OCRTOC_software_package}.
\end{abstract}

\section{Introduction}
Robotic grasping and manipulation has long been a research focus for the robotics and computer vision communities. In recent years, significant progress has been made in various related fields, such as robotic grasping \cite{fang2020graspnet,mahler2018}, object pose estimation \cite{xiang2018,wang2019dense}, motion planning for manipulation \cite{wang2020manipulation,ichnowski2020gomp} and control \cite{Bouyarmane2018quadratic,wang2019impact}. However, establishing fair performance assessment tools and quantifying the progress in the robotic manipulation domain remain as a challenge to date.
In the last decade, the robotics community tackled this challenge via several competitions (e.g. \cite{sun2016robotic,eppner2016lessons}) and benchmarking protocols (e.g. \cite{bekiroglu2019benchmarking,morgan2019benchmarking,falco2018performance}). While significant progress has been made in reproducible experimentation and performance assessment, the results are platform dependent, i.e. the experiments are executed on different platforms and therefore the algorithmic performance cannot be assessed in isolation. 

\begin{figure}[t]
	\centering
	\includegraphics[width=0.6\linewidth]{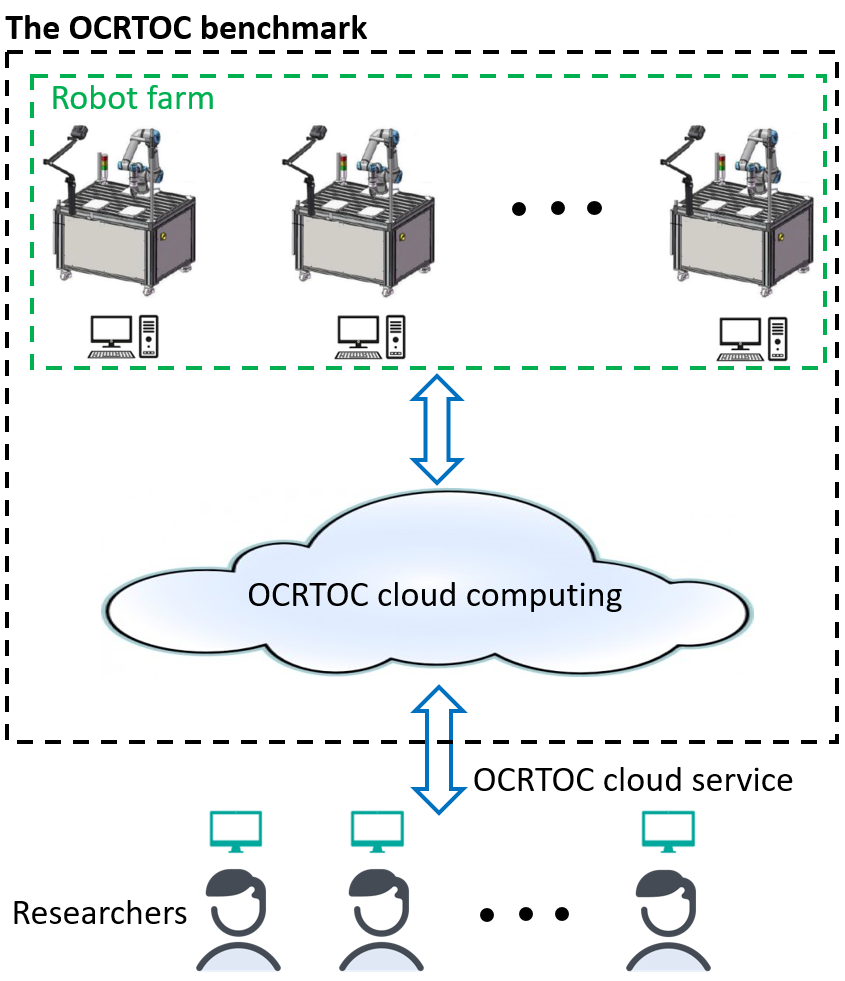}
	\caption{System overview of the OCRTOC benchmark.}
	\label{fig:ocrtoc_overview}
\end{figure}

Taking all these aspects into account, we propose a cloud-based benchmark system for robotic grasping and manipulation, called \textbf{the OCRTOC benchmark}. OCRTOC is the abbreviation of \textit{Open Cloud Robot Table Organization Challenge}. This benchmark focuses on object rearrangement scenarios. Object rearrangement is a canonical robotic task, and it embodies many challenging robotic aspects such as object grasping, recognition, manipulation planning and reasoning~\cite{rearrangement2020}. We facilitate a remote robot farm with several identical robotic setups, performing table organization tasks with varying levels of difficulty. As depicted in Fig.~\ref{fig:ocrtoc_overview}, researchers upload their solution through the OCRTOC cloud service, where their solution is evaluated both in simulation (using cloud computing) and on real robot hardware from the robot farm. The required software tools for developing and uploading solutions to our real robot platforms and the simulation environment to test them are publicly available to researchers. The tasks and evaluation metrics are standardized and are applied in the same way for all users. Therefore, we perform fair comparison between uploaded solutions  developed by different researchers, since all the algorithms undergo exactly the same experiment conditions. This scheme also allows researchers to conduct manipulation research without having their own robot hardware,  therefore lowering the barriers for robotics research. We believe that such an approach will make it easier to identify the shortcomings of the state-of-the-art methods, and promote the development of new algorithms to increase the robustness of robotic manipulation solutions. 

The OCRTOC benchmark is worldwide one of the first cloud-based benchmarks for robotic grasping and manipulation that provides a software system, standardized tasks, and evaluation with both simulation and real robot hardware. To show the effectiveness of the OCRTOC benchmark, we held a competition in conjunction with IROS 2020. In total, 59 teams from five continents took part in this competition, whose results are presented here. We plan to organize our next competition in 2021 based on the lessons learned from the first edition.
\section{Related Work}
\label{sec:rel}
While various robotics-related domains, such as object recognition and segmentation, enjoy benchmarks for performance evaluation and comparison \cite{xu2018youtube,grenzdorffer2020ycb}, developing benchmarks for robotic manipulation has long been a challenge. At the core of this challenge is the contact-rich nature of the manipulation tasks, preventing the researchers to accurately represent manipulation operations in simulations \cite{calli2017}. This problem is tackled in the community by defining benchmarking protocols, proposing standardized hardware, and organizing competitions.

Benchmarking in robotic manipulation has gained a significant momentum in the last decade. The YCB Benchmarking Project \cite{calli2017} facilitated a common set of objects and various guidelines for developing experiment-based protocols and benchmarks. Several benchmarks are defined for robotic grasping \cite{bekiroglu2019benchmarking,bottarel2020graspa}, bi-manual manipulation \cite{garcia2020benchmarking,chatzilygeroudis2020benchmark}, and motion planning \cite{lagriffoul2018platform}. 3D printed and sensorized objects are also made available for benchmarking purposes in \cite{gao2020modular}. 

These benchmarks provide detailed descriptions of the experimental setup and process, and often times give the users the freedom to utilize any robotic platform available to them. While this freedom enables the benchmarks to be utilized by a large group of researchers, the differences in robotic platforms prevents a direct, one-to-one comparison between manipulation algorithms. Nevertheless, most of these benchmarks are carefully designed to minimize the effect of the hardware on the assessment metrics (e.g \cite{bekiroglu2019benchmarking}). We believe that these benchmarks are a significant step towards performance assessment and comparison in robotics.

In order to solely focus on algorithm comparison, common robotic platforms are also proposed \cite{yang2019replab, ahn2020robel,locobot}. An important advantage of such platforms is that they permit continuous evaluation. Assuming that the research labs purchase these relatively inexpensive platforms and follow the same experimental procedures, a more direct quantitative performance comparison can be achieved. These platforms can also be utilized together with the aforementioned benchmarks to address the issues regarding cross-platform comparison.

Perhaps, the most system-level focused approach for assessing manipulation performance is the robotics competitions. In DARPA Robotics Challenge \cite{krotkov2017darpa}, teams competed with humanoid platforms for accomplishing various tasks such as removing debris, opening doors, closing a valve and using tools to break through a concrete panel. The Amazon Picking Challenge \cite{eppner2016lessons} was focused on bin/shelf picking tasks, geared towards warehouse robotics. The Robotic Grasping and Manipulation Competition \cite{sun2016robotic} is being organized since 2016. This competition has various tracks, such as service robotics, manufacturing and logistics, each of which requires teams to successfully execute complex manipulation operations such as picking objects from stacked configuration, using tools to achieve a task (e.g. using a tong to pick a ice cube), inserting objects with tight tolerances. In these competitions, teams build and utilize their own robotic setup. The competitions provide excellent opportunities to assess the performance of the state-of-the-art solutions and trigger innovations, resulting in creative approaches that help solve the target tasks. Generally, performance comparisons can be achieved in the system level, and the solutions are highly engineered towards the target tasks and conditions. 

Our benchmark and the associated competition are cloud-based, allowing users to utilize a common platform remotely and utilize the same experimental procedure. Such an approach allows us to address the cross-platform comparison issues and provide a common and unified way of establishing quantified assessments. There are a small number of cloud-based benchmarking tools currently available. Real Robot Challenge \cite{real-robot-challenge} provides a platform for in-hand manipulation research, and was used in 2020 for a robotic competition. The Robotarium \cite{pickem2017robotarium} is a remote experimentation setup that focuses on swarm robotics. Also, the KUKA Robot Learning lab at KIT \cite{kitrrl} offers the option to access a robotic lab with industrial robots over the internet. RoboTurk \cite{mandlekar2018roboturk} is a cloud-based platform that allows users to collect task demonstrations through teleoperation. To the best of our knowledge, our benchmarking platform and the corresponding challenge provide the only remote and cloud-based scheme that focuses on robotic grasping and object rearrangement.

The complexity of a rearrangement task depends on multiple factors, including object features, goal definition (exact or coarse rearrangement), motions that are needed for the rearrangement (random or ordered), and fidelity (real or simulation). We list the main differences of some existing rearrangement tasks in Table \ref{table:task_taxonomy}:
\begin{enumerate}[label=(\Alph*)]
	\item Amazon Picking Challenge~\cite{eppner2016lessons}.
	\item AI2-THOR Rearrangement~\cite{AI2-THORRearrangement}.
	\item Transport Challenge with ThreeDWorld~\cite{TDW-Transport}.
	\item RLBench~\cite{rlbench2020}.
	\item RGMC 2020~\cite{competition_iros2020} - Service Robot Track.
	\item RGMC 2020~\cite{competition_iros2020} - Manufacturing Track.
	\item OCRTOC.
	
\end{enumerate}
\begin{table}[ht]
	\centering
	\small
	\caption{Differences of some existing rearrangement tasks}
	\resizebox{\linewidth}{!}
	{
		\begin{tabular}{|c|ccc|c|c|c|}
			\hline
			tasks& \multicolumn{3}{c|}{object features}  &goal &motion & fidelity \\\hline
			(A)&deformable& single-body& unknown  &coarse& random &real\\
			(B)&rigid& articulated& partially known  &exact& ordered &sim\\
			(C)&rigid& single-body& known  & coarse& random&sim\\
			(D)&rigid& articulated& partially known&exact& ordered &sim\\
			(E)&deformable& single-body& known &coarse& ordered &real\\
			(F)&deformable& single-body& known &exact& ordered &real\\
			(G)&rigid& single-body& partially known  & exact& ordered&real \& sim\\
			\hline
		\end{tabular}
	}
	\label{table:task_taxonomy}
\end{table}

\section{The OCRTOC Benchmark System}
\label{sec:method}

\subsection{Task definition}
In the OCRTOC benchmark system, we focus on the task of table organization. The objects are selected from a subset of objects that can be easily found in the daily life. For our task, organization means that  objects should be placed in a specific physical state that is represented by a pose (position and orientation) defined in a certain coordinate system. Imagine the case that a service robot needs to set up the tableware for dinner: simple picking and placement without considering the final object pose will not fulfill the diner's wish of having nicely organized tableware, e.g. the cutlery, which normally contains multiple pieces, should be placed in a specific configuration.

A particular rearrangement task can be specified as $\mathbb{X}$:
\begin{equation}
	\mathbb{X}:=\{\mathbb{O},C^I, C^G\}
	\label{equ:tasks_definition}
\end{equation}
where $\mathbb{O}$ indicates the set of involved objects, $C^I$ and $C^G$ are the initial and target configuration. 
A scene configuration describes the physical state of the involved objects. There are many ways for representing such configurations. For instance, a configuration can be represented by an array, with each entry describing the 6D pose of the corresponding object. Scene configurations can also be represented by point clouds, or reference images, where 6D poses of individual objects can be extracted. 

\begin{figure}
	\centering
	\includegraphics[width=0.8\linewidth]{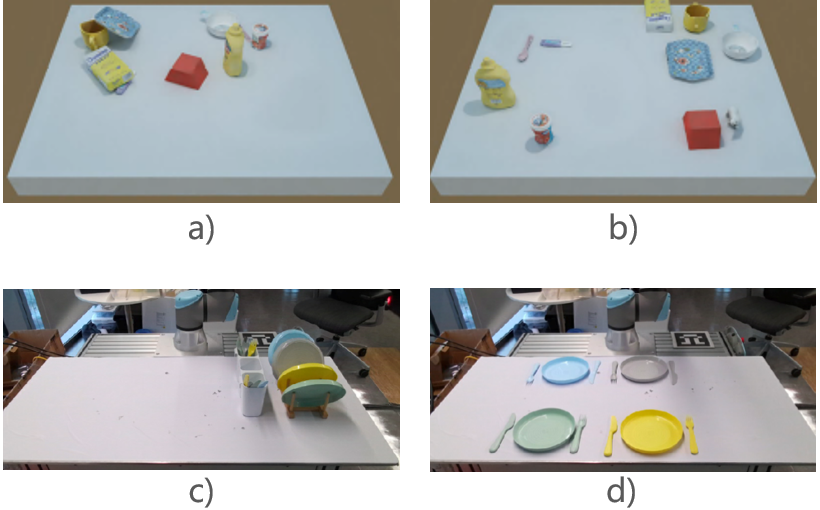}
	\caption{Examples of rearrangement tasks:  Initial scene a) and  target scene b) of a simulation task, and initial scene c) and target scene d) of a real robot task.}
	\label{fig:task_definition}
\end{figure}

The OCRTOC benchmark contains a list of predefined tasks $\mathbb{X}_i, i =1 \ldots N$. In each task, the contestant's solution needs to organize objects according to the target configuration so that the initial scene is transformed into the target scene after the manipulation. 
Two example tasks are illustrated in Fig. \ref{fig:task_definition}. 

\subsection{Performance evaluation}
The result of a contestant's solution is called the solution scene. The performance of a solution is evaluated based on the similarity between the solution scene and the target scene, i.e. how precisely the objects are placed compared to the target configuration. For each object, we compute the 3D Euclidean distance between its pose in the solution scene and its pose in the target scene. 
For each object's distance error, we set an upper bound that is related to its size, to avoid penalizing a single object too much. The error of each task is the average distance error of all involved objects. The performance of a solution is ranked according to its averaged error of all tasks. In case of the same distance error, the team with less execution time will be ranked higher.

\begin{algorithm}
	\caption{Performance evaluation of a single task}\label{algo:task_eval}
	\begin{algorithmic}[1]
	\Require{
		 Configurations of solution scenes $C^S$ and configurations of target scenes $C^G$}
	\Ensure{The 3D Euclidean distance error averaged over all objects.} 
	
	\Function{EvaluatePerformance}{$C^S,C^G$}
	\State $E=0$
	\For{target object $o_j$, with $j=1,...,M$}
	\State $ E +=\min(\textbf{EDE}(o_j,C^G,C^S),
	\textbf{UEB}(o_j))$
	\EndFor
	\State $E /= M$, with $M$ being the number of objects.
    \Return{$E$} 
	\EndFunction
	\end{algorithmic}
\end{algorithm}

The overall process of performance evaluation of a single task is explained in Algorithm \ref{algo:task_eval}. Here, \textbf{UEB}$(o_j)$ is the upper bound error of the object $o_j$. Using this function, the upper error bound of each object is set to a fixed value. In this way, the performance of a solution is not severely affected by some poorly manipulated objects, such as objects knocked away by the manipulator. In the OCRTOC 2020 competition, \textbf{UEB}$(o_j)$ was set to a value that is relevant to the object size:

\begin{equation}
	\textbf{UEB}(o_j) = 5\times(L_j + W_j + H_j)/3,
	\label{equ:upper_bound}
\end{equation}
where $L_j, W_j, H_j$ are the length, width and height of the object model bounding box for $o_j$ respectively. Starting from the OCRTOC 2021 competition, \textbf{UEB}$(o_j)$ will be set to the same constant value that is irrelevant to the object size.

\begin{figure}
	\centering
	\includegraphics[width=0.8\linewidth]{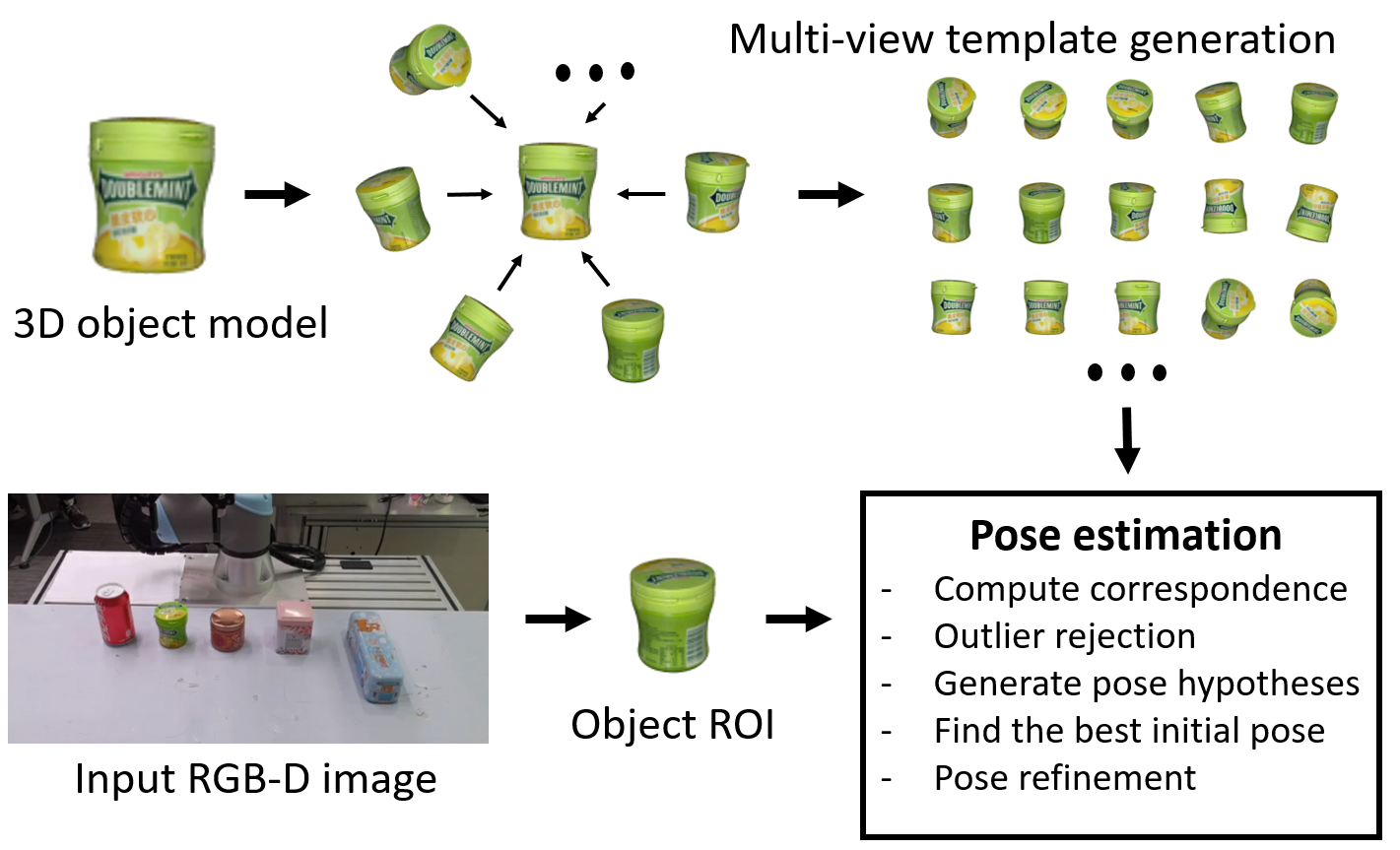}
	\caption{The general idea of the applied method for object pose estimation: 1) Generate enough templates: For each object, a lot of images are rendered using the object's model from different viewpoints. 2) Get the object bounding box from the input image. 3) Estimate the object pose. This step contains several sub-steps, such as finding correspondence between the input image and the generated object templates, outlier rejection, generation of pose hypotheses based on a PnP method, finding the best initial pose, and performing pose refinement if needed.}
	\label{fig:template_based_pose}
\end{figure}

The function \textbf{EDE}$(o_j,C^G,C^S)$ calculates the 3D Euclidean distance error for the object $o_j$. We apply a similar metric as defined in \cite{Hinterstoisser2012} to realize this function:
\begin{equation}
\resizebox{.9\hsize}{!}{$\textbf{EDE}(o_j,C^G,C^S) = \mathop{avg}\limits_{\mathbf{p}\in \mathbf{B}_{o_j}}||(\mathbf{R}^G_i\mathbf{p}+\mathbf{T}_i^G)-(\mathbf{R}_i^S\mathbf{p}+\mathbf{T}_i^S)||$},
	\label{equ:3d_distance}
\end{equation}
where $\mathbf{B}_{o_j}$ is a manually defined cube that is aligned with the axes of the object model $o_j$. The edge length of  $\mathbf{B}_{o_j}$ is equal to $(L_j + W_j + H_j)/3$. $\mathbf{p}$ indicates the eight vertices of $\mathbf{B}_{o_j}$. $\mathbf{R}_i^G$ and $\mathbf{T}_i^G$ are the rotation and translation defined in the target configuration $C^G$. $\mathbf{R}_i^S$ and $\mathbf{T}_i^S$ indicate the estimated rotation and translation in the solution configuration $C^S$. $\mathbf{R}_i^S$ and $\mathbf{T}_i^S$ are calculated in a similar way as described in \cite{Hinterstoisser2012}. The method for object pose estimation is illustrated in Fig.~\ref{fig:template_based_pose}.

\section{Implementation}
\label{sec:implementation}
\subsection{Software system}
\begin{figure}[ht]
	\centering
	\includegraphics[width=0.8\linewidth]{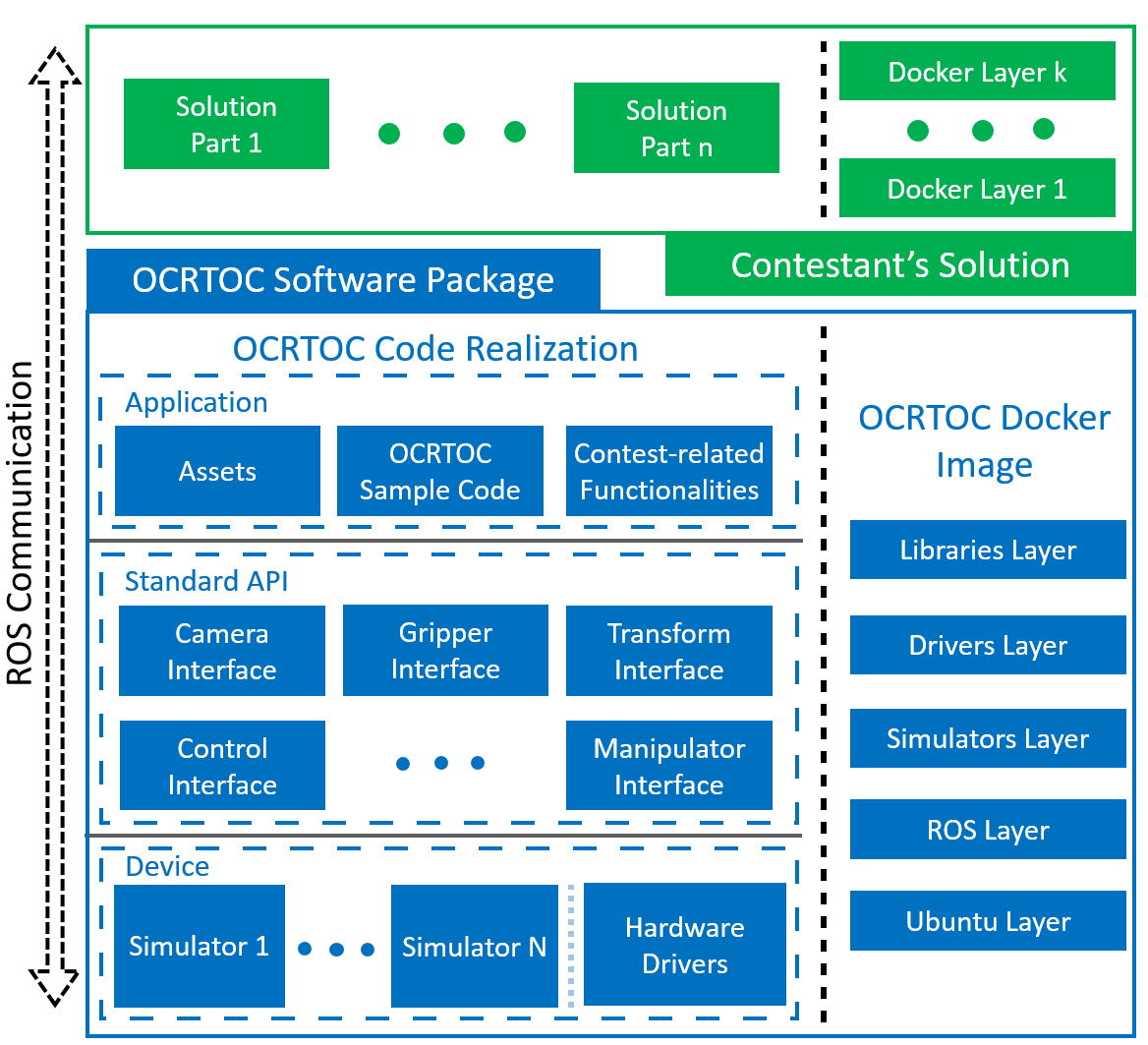}
	\caption{Software architecture of the OCRTOC system.}
	\label{fig:Software_architecture}
\end{figure}

The software architecture of the OCRTOC benchmark system is illustrated in Fig. \ref{fig:Software_architecture}. The \emph{OCRTOC Software Package} (shown in blue) is developed based on the Docker technology \cite{docker2019} and serves as the base system of OCRTOC. The \emph{Contestant's Solution} (shown in green) is the part that contestants need to develop on top of the OCRTOC software package. Data communication between both parts is realized using ROS mechanisms. Docker is an OS-level virtualization technology and makes software development independent from the underlying operation system. It guarantees that Contestant's Solution delivers the same runtime performance in both contestant's local machine and the OCRTOC benchmark system. In this way, we provide contestants a highly efficient development environment, so that they can concentrate on developing algorithms and solving the OCRTOC tasks.

OCRTOC Software Package contains two parts: \emph{OCRTOC Docker Image} and \emph{OCRTOC Code  Realization}. In OCRTOC Docker Image, several layers of dependencies are already pre-installed, such as the Ubuntu Layer (Ubuntu 18.04), and the ROS Layer (ROS melodic \cite{ros-melodic}). If contestants need other dependencies, they can extend OCRTOC Docker Image by installing their own dependencies in Contestant's Solution. OCRTOC Code Realization is the main part of OCRTOC Software Package and contains three layers:

\begin{itemize}
	\item Device: Drivers of hardware components or their counterparts in simulators are provided to enable the use of devices such as sensors and manipulators. In the current version, we support two simulator types, namely  PyBullet\cite{coumans2019} and SAPIEN\cite{Xiang_2020_SAPIEN}.
	\item Standard API: Software APIs for various functionalities are provided to ease the development of Contestant's Solution, including a camera interface, a control interface, or the interface to get the calibration information. 
	\item Application: Application-level software components include assets, OCRTOC sample code, and contest-related functionalities. Assets indicate non-programmable software artifacts such as object models and OCRTOC tasks. OCRTOC sample code demonstrates how to use the whole OCRTOC package. Contestants can easily build their solution by incorporating their solution parts into the sample code. Contest-related functionalities are mainly used in the evaluation of Contestant's Solution.
\end{itemize}

\subsection{Simulator}

\subsubsection{Object Models}
\begin{figure}[ht]
	\centering
	\includegraphics[width=0.6\linewidth]{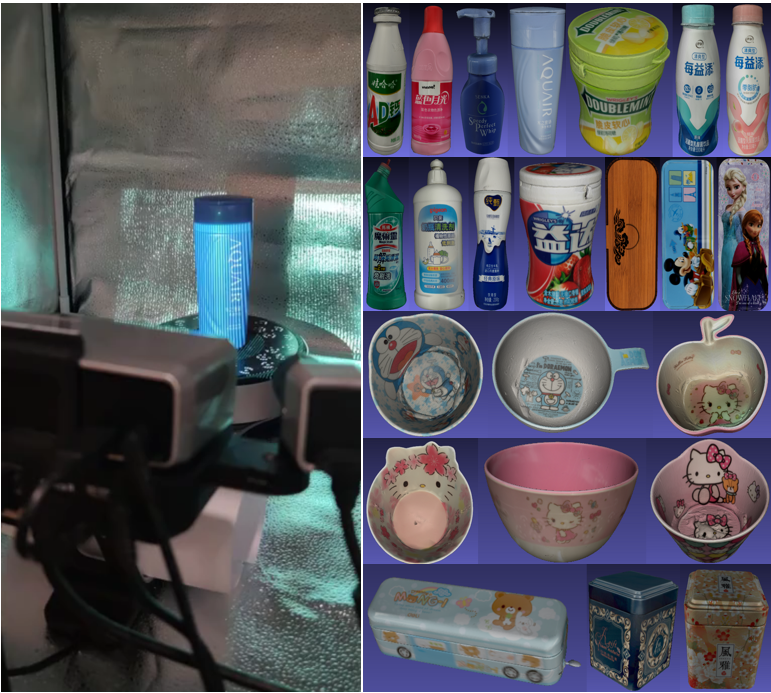}
	\caption{The scanning process (left), and some scanned object models. (right)}
	\label{fig:objects}
\end{figure}

In the current version, the object database contains 154 object models (3D triangle mesh models with texture). Among them, 53 are selected from the YCB dataset \cite{calli2017}. The other 101 object models are scanned using a 3D scanning device\cite{shining3d}. The information about the objects in the database is given in table \ref{table:objects_list}. An example of the scanning process and some examples of object models are depicted in Fig. \ref{fig:objects}.

\begin{table}[ht]
	\centering
	\small
	\caption{Types and number of object models}
	\begin{tabular}{|ll|ll|ll|}
		\hline
		Type & Quantity & Type & Quantity & Type & Quantity \\\hline
		Toy  & 25 &Plate  & 9 &Cup  & 24\\
		Cutlery  & 8 &Box  & 21 &Scissor  & 1\\
		Bowl  & 17 &Kettle  & 1 &Can  & 16\\
		Pen  & 2 &Bottle  & 15 & Tool &15 \\

	 \hline
	\end{tabular}
	\label{table:objects_list}
\end{table}

In general, object models are always available for contestants during the runtime of their solution. Corresponding model data, such as triangle mesh and texture data, is accessible using predefined APIs. However we divide the total number of objects into a trial set and a contest set. The object models in the trial set can be downloaded for training purposes. The object models in the contest set are only available during online evaluation in our benchmark system. Tasks that should be solved in the evaluation contains objects from both sets. In this way we aim to test the generalization ability of the contestant's solution. 

\subsubsection{Generation of simulation tasks}
\begin{figure}[ht]
	\centering
	\includegraphics[width=0.70\linewidth]{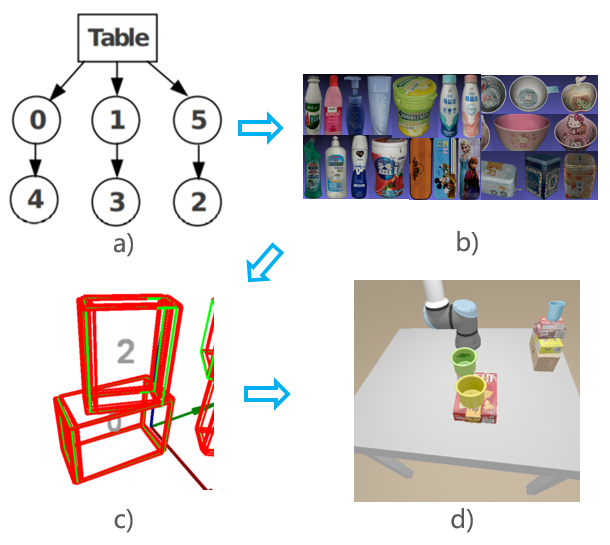}
	\caption{Procedure for scene generation. a) A scene graph describing the construction of a scene \cite{LIU2015110}. b) Instantiate objects through sampling in the object database. c) Instantiate object poses while keeping the inter-object relationship. d) A generated scene that passed the validity check.}
	\label{fig:scene_generation}
\end{figure}

We use a scene graph inspired by the description proposed in \cite{LIU2015110} to model scenes in simulation. The scene graph models the inter-object relationships of the underlying objects. In the scene graph, each object is assigned with an initial 6D pose. Based on the same scene graph, different scenes can be generated through sampling in the object database and by adding relative transformations to the initial pose. The generated scenes are then validated by a physics engine, so that invalid scenes are removed. The generated tasks can be run in all simulators that are supported by our Benchmark system. In all, we generated 1400 simulation tasks. The trial set with 1100 tasks can be downloaded for training purposes. The contest set with 300 contest tasks is only available in our online benchmark system. The general procedure of scene generation is depicted in Fig. \ref{fig:scene_generation}.

\section{OCRTOC Competition 2020}
\label{sec:evaluation}
\subsection{Competition format}
The first OCRTOC competition was held in conjunction with IROS 2020.
The competition contained two stages: a simulation and a real robot stage. Each stage began with a trial period and ended with a contest. In total, 59 teams from all over the world took part in this competition, coming from five continents: 44 from Asia, 9 from North America, 3 from Australasia, 2 from Europe, 1 from South America. Detailed information about participating teams can be found on the OCRTOC homepage \cite{ocrtoc-home}.

\textbf{The simulation stage} (17.Aug.2020 - 05.Oct.2020):
In the trial period of the simulation stage, contestants developed their solution based on the trial task set. In this period they could try out the OCRTOC Software Package and get familiar with the OCRTOC benchmark system. In the simulation contest, contestants uploaded their solution to the online evaluation system of OCRTOC, where their solution was evaluated using contest tasks. After the submission deadline, contestants were ranked by their performance, and those whose performance was better than the predefined baseline qualified for the real robot stage. 

\begin{figure}[ht]
	\centering
	\includegraphics[width=\linewidth]{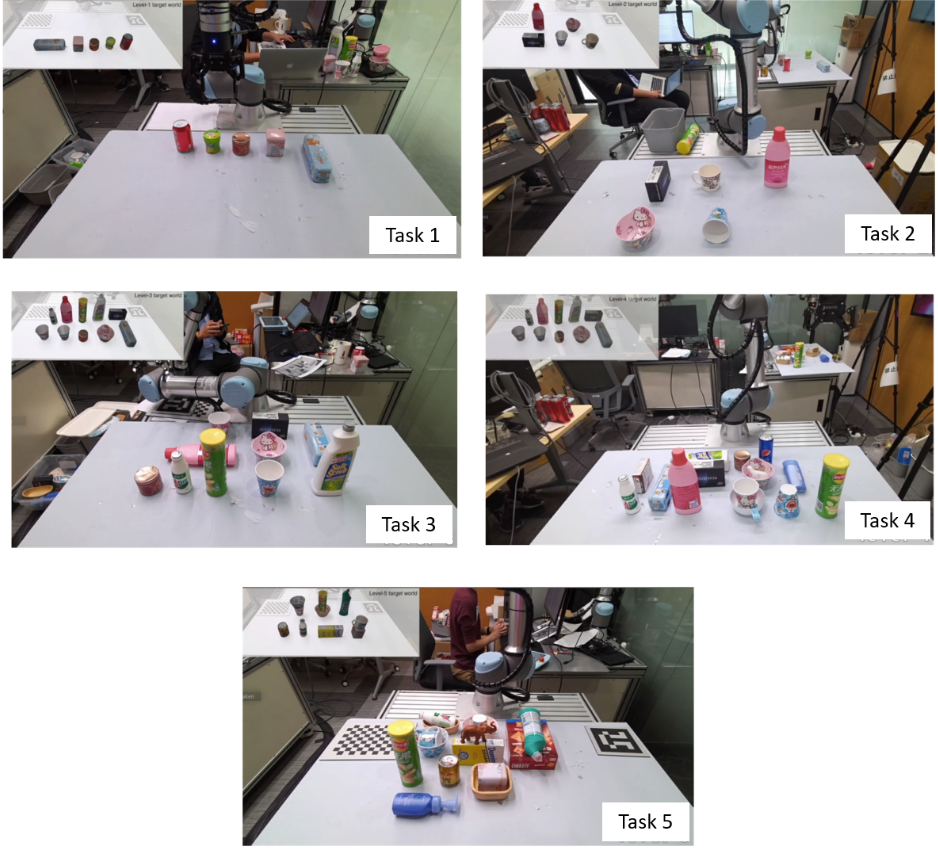}
	\caption{Tasks used in the real robot contest. The target scene of each task is shown in the upper left corner of each sub-figure. Each contestant's solution was evaluated three times using these five tasks, and the best performance among the three runs was taken as the final performance for the corresponding contestant.}
	\label{fig:real_tasks}
\end{figure}

\textbf{The real robot stage} (11.Oct.2020 - 06.Nov.2020):
In the trial period of the real robot stage contestants could book trial sessions from the organizer to test their solution on real robot hardware. Each team had a limited time period for trials. During trials, most of the teams collected data for algorithm training. In the contest their solution was evaluated using the five contest tasks shown in Fig. \ref{fig:real_tasks}. 

\subsection{Hardware setup}
\label{sec:hardware2020}
The hardware setup in the real robot stage included:
\begin{itemize}
	\item A stationary camera: Kinect DK camera.
	\item A wrist-mounted camera: RealSense D435i.
	\item A manipulator: UR5e from Universal Robots.
	\item An end-effector: 2-finger gripper 2F-85 from Robotiq.
	\item A PC as the computing device: CPU Intel Xeon E-2246G, Memory 32GB DDR4, GPU Nvidia Geforce RTX2080 with 8GB memory.
\end{itemize}

\subsection{Definition of baseline performance}
Each task has a default error once its initial scene and target scene are defined. In the OCRTOC 2020 competition, a task's default error was defined as the sum of the equation $\textbf{UEB}(o_j)$ (eq.~\eqref{equ:upper_bound}) over all objects in the task. The default error of a task was considered as the baseline performance, and we measured how much improvement (in percentage) each team achieved using their solution.

\subsection{Results of the simulation contest}
In the simulation stage, contestants' solutions were evaluated using 100 contest tasks using both simulators. For each team, the better result from the two simulators was considered as the team's final performance. The evaluation was executed using the Elastic Compute Service of Alibaba Cloud \cite{ecs-ali}. In total, 14 teams passed the simulation contest, and their results are given in table \ref{table:sim_reuslt}. 

\begin{table}[ht]
	\centering
	\small
	\caption{Results of the simulation contest. Only teams that beat the baseline are listed. Error is given in centimeter.}
	\begin{tabular}{|l|ll|l|ll|}
		\hline
		Rank  & Error & Improvement &Rank  & Error & Improvement \\\hline
		1    & 29.21 &10.3\% &8    & 31.82 &  2.3\% \\
		2    & 29.43&  9.6\% &9    & 31.84 & 2.2\%\\
		3    &   30.15&  7.4\%& 10    & 31.84& 2.2\%\\
		4    & 30.28 &  7.0\% &11    & 31.88& 2.1\% \\
		5    & 30.56 &  6.2\% &12    & 31.88 &  2.1\%\\
		6    &30.95 &  4.9\% &13    & 32.15  &2.0\%\\
		7    & 31.04  &  4.7\%&14    &32.22  & 1.0\% \\\hline
		
	\end{tabular}
	\label{table:sim_reuslt}
\end{table}

In the simulation stage, contestants had to perform grasping. Moving objects using magnetic force (like a suction cup) was not allowed. In simulation tasks, the ground truth of object poses were used to calculate the performance (as described in Algorithm \ref*{algo:task_eval}). Improvement refers to the performance improvement (in percentage) that was realized by the corresponding solution compared with the baseline performance. All the teams that achieved performance improvement against the baseline qualified for the real robot stage.

\subsection{Results of the real robot contest}
In the real robot stage, contestants' solutions were evaluated using the five tasks shown in Fig.~\ref{fig:real_tasks}. For each contestant, the evaluation was run three times, and the best performance among the three runs was taken as the final performance for this contestant. The execution time for each task was limited to 10 minutes. The OCRTOC team coordinated the executions with the contestants, and was responsible for resetting the table layout after each execution. In case of hazards, such as breaking things and safety issues, the task execution was terminated. To make the evaluation as fair as possible, we made task-specific markings on the table, so that the initial configuration of the used tasks could be repeated. 

The results of the real robot contest are given in Table \ref{table:real_reuslt}. As the task difficulty increased from Task 1 to Task 5, the performance declined accordingly with increasing error (see \textbf{Average task error of top five teams}). The best ranked team could only achieve an improvement of 31.1\% (averaged over all five tasks) compared to baseline, even though they had a relatively high grasp success rate of 78.8\%. From the 14 teams that qualified for the real robot stage, only five of them could achieve improvement compared to the baseline. 

As the scenes (either initial or target) became more complex, the performance of the contestants declined rapidly. The best performance among all teams achieved a relatively good improvement in Task 1 (61.3\%) and Task 2 (47.5\%). However their scores declined to about 30\% in Task 3, 20\% in Task 4, and 16\% in Task 5. This was partially due to the increasing complexity for grasping and pose estimation in cluttered scenes. Moreover, from our observation, none of the teams developed a reasonable motion planning module that was good enough to plan motion sequences in highly cluttered scenes. These results indicate that table organization is still a challenging robotics task to solve.

\begin{table*}[ht]
	\centering
	\small
	\caption{Results of the real robot contest. Only teams that beat the baseline are listed. Errors are given in centimeter. Grasp success rate is defined as the ratio between successfully performed grasps and the total number of attempted grasps. T1 to T5 refer to Task 1 to Task 5, as shown in Fig. \ref{fig:real_tasks}. The improvement in percentage for T1 to T5 is given in brackets. The best performance in each task is highlighted in blue.}
	\resizebox{\textwidth}{!}
	{
		
	\begin{tabular}{|l|lll|lllll|}
		\hline
		Rank & Error & Improvement & Grasp success rate &T1 Error & T2 Error & T3 Error & T4 Error & T5 Error\\\hline
		1  & 34.29 & 31.1\%	&26/33=78.8\% &19.29 (53.5\%) & \textcolor{blue}{27.59 (47.5\%)} & 41.29 (21.2\%)& \textcolor{blue}{41.62 (20.6\%)}& \textcolor{blue}{41.64 (16.5\%)} \\
		2  & 35.02 &  29.6\%	&21/34=61.8\% &\textcolor{blue}{16.07 (61.3\%)}& 33.99 (35.4\%)& \textcolor{blue}{36.44 (30.5\%)}& 42.87 (18.2\%)& 45.73  (8.3\%)\\
		3  & 41.06 &  17.5\%	&13/17=76.5\% &29.78 (28.2\%)& 38.90 (26.0\%)& 43.68 (16.7\%)& 43.08 (17.8\%)& 49.84 (0.0\%) \\ 
		4  & 42.26 &  15.1\%	&9/32=28.1\% &34.02 (18.0\%)& 40.22 (23.5\%)& 43.79 (16.5\%)& 46.93 (10.5\%)& 46.34  (7.0\%)\\
		5  & 46.47 &  6.6\%	&2/3=66.7\% &25.08 (39.6\%)& 52.59 (0.0\%)& 52.41 (0.0\%)& 52.41 (0.0\%)& 49.84 (0.0\%) \\
		\hline
		Baseline  &49.75  & N/A	&N/A &41.49 & 52.59 & 52.41 & 52.41 & 49.84  \\\hline
		\multicolumn{4}{|c|}{\textbf{Average task error of top 5 teams}}&24.85 (40.1\%)& 38.66 (26.5\%)& 43.52 (17.0\%)& 45.38 (13.4\%)& 46.68  (6.3\%)\\\hline
	\end{tabular}
	}
	\label{table:real_reuslt}
\end{table*}

\subsection{Summary of leading strategies}
We analyzed the methods used by the top three teams and came to the following conclusion.

They all used a multi-step pipeline to solve the rearrangement problem. In each step they tried to tackle a sub-problem, such as pose estimation, grasping, or planning. None of them treated the rearrangement problem as a pure end-to-end deep learning problem. Their general pipelines were quite similar.  Normally, it started with object pose estimation, and then grasp poses were generated. After that a planning module was applied to generate motion for the robot. 

For object pose estimation, both traditional methods and deep learning methods were used by top teams. Examples of traditional methods included 2D feature matching methods, 3D distance clustering methods, and 3D point cloud ICP methods. For deep learning methods, synthetic data were first used to pre-train a neutral network, which was then enhanced by manually labeled real-world data. Grasp pose generation was done mostly based on the estimated object pose and the given object model. For motion planning, some basic heuristic rules were used to implement a planning strategy. In easy tasks, these rules worked well, such as task 1 and task 2. However, they failed dramatically in difficult tasks with highly cluttered scenes, such as task 3 to task 5.

\section{OCRTOC Competition 2021}
\label{sec:conclude}
We have improved the OCRTOC benchmark and will continue the competition in 2021. The main changes introduced with respect to the first edition are listed below.

\subsection{Competition format}
The competition in 2020 was a one-shot action with a duration of two and a half months. Many potential teams could not participate due to schedule conflicts with their daily work. We decided to change the competition format to allow monthly submissions, allowed during the whole year. Winners will be announced by the end of each year. In this way, teams can flexibly plan their participation. For 2021, the competition starts in July.

\subsection{Baseline solution and training data}
In the 2020 edition, we noticed that it was a tedious job for contestants to build a complete solution from scratch. In addition, many teams needed to collect training data during real robot trials and labeled the collected data by themselves. For the 2021 competition  we will provide a baseline solution, which can be used by contestants to build their own solution. This should enable more contestants to join the competition. Our baseline solution contains mainly the following modules:
\begin{itemize}
	\item Planning: A high-level motion planning module is provided based on PDDLStream \cite{garrett2020pddlstream} that plans motion sequences in the symbolic domain. Given the motion sequences, common trajectory planners can be used to generate the final trajectories that need be executed by the robot.
	\item Perception:
	\begin{itemize}
		\item Grasping: An end-to-end grasping method \cite{fang2020graspnet} based on point cloud is provided to generate grasp poses.
		\item Object pose estimation: Two methods are provided for pose estimation. One is based on keypoint matching on the image plane, and the other one is based on point cloud matching.
	\end{itemize}
\end{itemize}
The provided baseline solution can solve the tasks to some extent. Contestants can start with this baseline solution and improve the performance gradually.

\begin{figure}[ht]
	\centering
	\includegraphics[width=\linewidth]{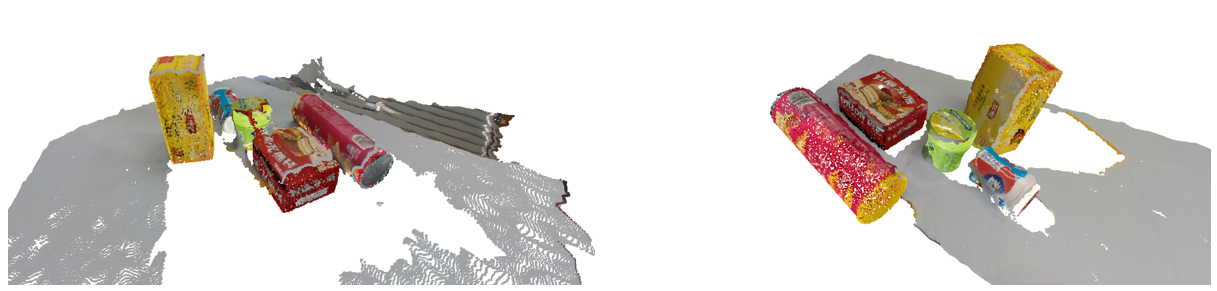}
	\caption{An example of a labeled scene: the same scene in two view angles. Object models are projected into the scene using the labeled per-scene poses.}
	\label{fig:dataset}
\end{figure}

Moreover, we provide a labeled real world dataset in the OCRTOC benchmark. In this dataset, scenes containing multiple objects are captured by a rgbd camera in 79 different view angles. Triangle mesh models with texture are provided for all involved objects. The per-scene 6D object poses are labeled. For each rgb image, per-frame object poses and 2D object masks are also provided. An example of this dataset is shown in Fig.~\ref{fig:dataset} We will extend this dataset continuously. Currently, statistics about the dataset is as follows: 
\begin{itemize}
	\item Number of objects: 33.
	\item Number of scenes: 76.
	\item Number of labeled rgbd images: 6004.
	\item Number of labeled per-frame poses: around 30000.
\end{itemize}

\subsection{Hardware update}
To open up more possibilities for robot control and motion planning, we will use 7-axis manipulators with force feedback in 2021, instead of the 6-axis robot used in 2020. The selected model is Panda from Franka Emika \cite{franka-panda}. The used end-effector is the parallel-jaw gripper that is delivered with Panda. The rest of the hardware setup is the same as in 2020 (section \ref{sec:hardware2020}).

\subsection{Tasks}
\begin{figure}[ht]
	\centering
	\includegraphics[width=\linewidth]{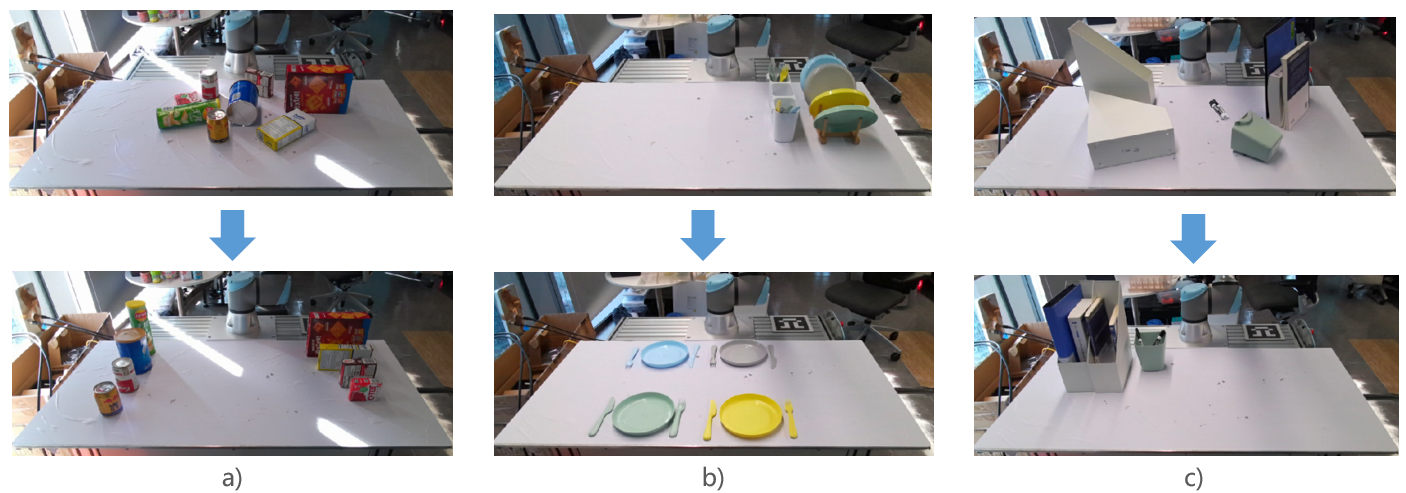}
	\caption{Three example tasks for OCRTOC competition 2021: a) food organization, b)  tableware organization, c) stationery organization. Initial scenes are shown in the upper row, and the corresponding target scenes are shown in the lower row.}
	\label{fig:task-2021}
\end{figure}

For the 2021 competition tasks will be more realistic, and they will align with use cases in our daily life. Three example tasks are depicted in Fig. \ref{fig:task-2021}, which are food organization, tableware organization and stationery organization.

\section{Conclusions}
\label{sec:conclusion}






In this paper, we proposed a cloud-based benchmark for robotic grasping and manipulation, called OCRTOC. This benchmark focused on the object rearrangement problem in the context of table organization. Our benchmark was facilitated with standardized software system, hardware setups, and tasks settings. Using this benchmark, algorithm performance of different teams could be evaluated on the same robot hardware in a fair way. In conjunction with IROS 2020, we held the first competition with the OCRTOC benchmark. Experimental results showed that rearrangement was still a challenging task, which required a whole functional skill set including at least pose estimation, grasping, and motion planning. In our experience, cloud-based benchmarks or competitions as such could greatly lower the barrier of conducting reproducible research. From all 59 teams, some did not have their own robot, and some were completely new to this field. Based on experience and lessons learned from the 2020 competition, we improved our benchmark in many aspects. We will continue the competition in 2021, tentatively, starting in June 2021.


\bibliographystyle{IEEEtran}
\bibliography{ocrtoc_reference}



\end{document}